\documentclass[12pt, letterpaper]{article}
\RequirePackage{latex_style}
\usepackage{caption}
\begin{document}


\title{\bf\color{titleblue}{A Time Series Graph Cut \\Image Segmentation Scheme\\ for Liver Tumors}}
\author{Laramie Paxton$^1$*, Yufeng Cao$^1$, Kevin R. Vixie$^1$, Yuan Wang$^1$, Brian Hobbs$^2$, Chaan Ng$^2$\\
  \small $^1$Department of Mathematics and Statistics, 
Washington State University\\
  \small PO Box 643113
Pullman, WA 99164, U.S.A.\\ \medskip
\small $^2$M.D. Anderson Cancer Center, 
University of Texas\\
\small 1515 Holcombe Blvd.
Houston, Texas 77030 \\ \medskip

  \small E-mail: *realtimemath@gmail.com
}


\date{}

\maketitle

\begin{abstract}
Tumor detection in biomedical imaging is a time-consuming process for medical professionals and is not without errors. Thus in recent decades, researchers have developed algorithmic techniques for image processing using a wide variety of mathematical methods, such as statistical modeling, variational techniques, and machine learning. In this paper, we propose a semi-automatic method for liver segmentation of 2D CT scans into three labels denoting \textit{healthy, vessel,} or \textit{tumor} tissue based on graph cuts. First, we create a feature vector for each pixel in a novel way that consists of the 59 intensity values in the time series data and propose a simplified perimeter cost term in the energy functional. We normalize the data and perimeter terms in the functional to expedite the graph cut without having to optimize the scaling parameter $\lambda$.  In place of a  training process, predetermined tissue means are computed based on sample regions identified    by expert radiologists. The proposed method also has the advantage of being relatively simple to implement computationally. It was evaluated against the ground truth on a clinical CT dataset of 10 tumors and yielded segmentations with a mean Dice similarity coefficient (DSC) of .77 and mean volume overlap error (VOE) of 36.7\%. The average processing time was 1.25 minutes per slice. 
\end{abstract}
\small{\tableofcontents} 
\section{Introduction}

Hepatic cancer is a common and increasingly deadly form of cancer worldwide. In the United States alone, the number of new cases increased 38\% from 2003 to 2012, with the number of deaths jumping 56\% \cite{us2010united}. Therefore, proper detection and diagnosis is imperative, as segmentation of healthy tissues and existing tumors allows for accurate staging and pre-surgical planning along with choosing the best treatment plan and providing postoperative care. The standard imaging technique used is multi-phase, computed tomography (CT) for its speed, resolution, and affordability, yet there is still a significant amount of variation present from the tumor type, pathological stage, contrast dose, and scan delay \cite{li2012new}. Moreover, the problem is made especially challenging due to the low contrast between tumor tissues and surrounding tissues and vessels, the high degree of variation among tumor shapes, and their ambiguous boundaries \cite{rajagopal2015survey}, \cite{wu20173d}. 

While clinicians have tended to rely on manual identification and segmentation by radiologists, this method is tedious, time-consuming, and carries a degree of variability among raters and thus lacks reproducibility \cite{kadoury2015metastatic}. Therefore, numerous interactive, semi-automatic, and automatic segmentation techniques have been developed in the past two decades. 

Methods such as clustering \cite{wu2007brain},  \cite{sinha2014efficient}, \cite{zhu2003automation}, thresholding \cite{kumar2016statistical}, \cite{chang2006survey}, and region growing \cite{Abdelsamea2014AnAS}, \cite{ganjre2014automated} are relatively simple to implement computationally, but rely only on pixel intensity and are thus subject to ``leakage'' on tumor boundaries \cite{wu20173d}. To overcome these limitations, researchers  have proposed using techniques such as adaptive region growing, adaptive thresholding, or combining clustering with other methods. For example, in \cite{li2011integrating}, a spatial fuzzy clustering approach is used with a level set method. The study in \cite{anter2013automatic} compares three semi-automatic region growing segmentation methods, and \cite{RajputASO} compares a region growing method with the Otsu method. In \cite{hore2016integrated}, a watershed algorithm was used with region growing in order to better  detect the initial seed points, and \cite{moghbel2016automatic} uses fuzzy c-means and random walkers algorithms.

Machine learning approaches \cite{zhang2018convolutional}, \cite{huang2014random},    \cite{li2015automatic}, \cite{kadoury2015metastatic}, \cite{ronneberger2015u}, include \cite{christ2017automatic}, which uses cascaded fully convolutional neural networks, and \cite{huang2013liver}, which represents voxels with a feature vector and trains the Extreme Learning Machine algorithm for voxel classification. While many promising machine learning techniques have been proposed in recent years, the drawbacks include the high computational requirements \cite{kadoury2015metastatic} and the large amount of high quality training data required during the training process \cite{wu20173d}.

Other popular methods include those that rely on energy minimization  to label different regions, such as active contour methods, including level set \cite{jin2011adaptive} and fast marching \cite{le2016liver} methods, and graph cuts. The active contour-based models, however, can often be too sensitive to the contour initialization due to the influence of local minima on the energy minimization and due to ``leaking'' on tumors with weak boundaries or when noise or poor contrast prevent the contour from stopping properly \cite{slabaugh2005graph}. Researchers in \cite{li2012new} have introduced a unified level set approach with fuzzy clustering to integrate image gradient, region competition, and prior information. Other efforts include \cite{li2011integrating}, mentioned above, and \cite{raj2016automated}, which uses a hybrid of Markov Random Field (MRF) level set, shape analysis, graph cut, and Support Vector Machines (SVM) classifier techniques. 

\subsubsection*{Graph Cuts}
Graph cuts \cite{ramyaimage}, \cite{boykov2006graph},   \cite{slabaugh2005graph}, \cite{vu2008image},   \cite{fang2011segmentation},  \cite{linguraru2012tumor},  on the other hand, differ from active contour methods in that they are most often not iterative and compute the global energy minimization. The also do not depend on initializations. Graph cuts are rooted in combinatorial optimization theory and seek to minimize an energy functional by making the minimum-cost cut that divides a connected, undirected graph into two disconnected pieces. The graph is typically represented with pixels as vertices and edges derived from 4- or 8-neighborhood connectivity among pixels called n-links. A source and sink node are added, and edges from each pixel to these two nodes are also made, called t-links  \cite{chang2014iterative}.  The energy functional contains both a data fidelity, or regional, term and a perimeter regularization, or boundary, term. The former penalizes based on classifying pixels in a certain region, foreground or background, while the latter penalizes based on pixel differences along boundaries or similar measures, with costs accruing only along those edges where cuts are made. The minimization is obtained based on the  \textit{max-flow min-cut theorem} \cite{dantzig2003max}, which states that the weight of the edges in the cut of minimum capacity in a flow network equals the maximal flow that can travel along the network, where the cut of minimum capacity represents the smallest overall weight of those edges that would disconnect the source from the sink if removed.

One of the most popular algorithms for computing the graph cut is the Boykov-Kolmogorov (BK) max-flow algorithm \cite{boykov2004experimental},  \cite{kohli2005efficiently},  \cite{kohli2008measuring}, which we utilize in the proposed model below. The main drawback with regular graph cut methods is their difficulty handling weak boundaries and noisy images, and so methods  have incorporated techniques to overcome this, such as the random walkers algorithm \cite{moghbel2016automatic}.
Other recent proposed models include \cite{wu20173d}, which utilizes a four-step process including a kernelized fuzzy c-means (FCM),  confidence connected region growing algorithm, and graph cut. In  \cite{chang2014iterative}, kernel density estimation is used to develop a nonlinear
statistical shape prior in such a way that the energy functional can be minimized through iterative graph cuts. And, \cite{stawiaski2008interactive} applies minimal surfaces and MRF's to the watershed transform to create an interactive graph cut method   using a region graph in place of a pixel graph. 

\subsection{Proposed Method}
In this paper, we propose a semi-automatic method of liver CT scan segmentation that incorporates the time series data for each pixel into a feature vector to make the BK algorithm more robust to weak boundaries and noise. This addition utilizes the differences over time in the intensities of the different tissue types as a result of their response to the injected contrast agent in each patient. Adding a simplified perimeter term and normalizing both terms in the energy functional, we perform the graph cut twice using the BK max-flow algorithm described above; once to separate the healthy and tumor tissues and again to separate the tumor and vessel tissues. Predetermined tissue means are provided in the functional from sample regions obtained from regions of interest (ROI's) produced by expert raters. To the best of our knowledge, this is the first time the time series data has been incorporated into 2D liver image segmentation in this way, and we show that doing so results in a relatively high degree of accuracy for a short computational time and an algorithm that is simple to implement computationally as compared to other proposed methods in the literature. For example, no training process is required.  


\section{Materials and Methods}
The proposed method was evaluated on a clinical CT dataset of 2D images containing 10 hepatic tumors obtained as dicom images from the M.D. Anderson Cancer Center at the University of Texas, and ground truth segmentations and liver masks were provided by expert raters for evaluation of the method along with ROI's for obtaining sample tissue means for healthy, tumor, and vessel tissues. While this was provided for 10 different slice depths for each tumor, we segmented only the most shallow depth so as to limit intensity variations from different body compositions as much as possible. In the case where more than one ROI was provided for a tissue type, we used the first one provided for tumor tissue and the second one provided for healthy tissue corresponding to the slide depth. We utilized phase 1 of a 64-stage CT scanner in which a total of 59 slices were present in each series taken 0.5 sec apart over 30 seconds. The pixel spacing was either .70 mm or .86 mm. The slice thickness was 5 mm, and the image resolution was 512 $\times$ 512 in all cases. All segmentations are performed on the 59th image in the series to allow the contrast agent to reach its full efficacy. The units of intensity in the CT images are Hounsfield units (HU). Note that all computations were performed using Matlab 2018a on a personal computer with 4 Gb of RAM and a 2.5 GHz Intel Core i5 CPU. 

\subsection{Preprocessing}

We use 59 images in each series taken over a sampling interval of 30 seconds. Let the image $I$ be the 59th image.   Utilizing the full time series, we form a feature vector for each pixel $p^i\in I$ using the values of the intensities at each time step for $p^i$. We do this without smoothing the images so that we preserve the time series intensity values for each pixel. We did not perform registration or motion correction on the series. Thus we have $$p^i_{59} \longrightarrow (p^i_1,p^i_2,p^i_3,\dots,p^i_{59}) \, \, \, \, \forall p^i\in I.$$

Since the three different tissue types, healthy, tumor, and vessel, each respond with varying intensities over the time series as the injected contrast agent is processed in the liver, we are able to 
 incorporate the temporal information in tissue intensity differences in addition to the spatial information in the initial image.

As a means of comparison with the proposed model, we also evaluate the data set using the BK graph cut on 1) the scalar pixel intensities after a $3\times 3$ neighborhood median filter is applied; and 2) the case in which each pixel is vectorized through the use of an averaging filter. In this case, we create a multiscale descriptor by taking the anisotropic average around each pixel at different length scales. That is,  we perform $10$ convolutions such that the first entry of each vectorized pixel, which we denote by $p^i_1$, is generated by taking the convolution over a $1\times1$ neighborhood of $p^i$. The second entry $p^i_2$ uses a 2 $\times $ 2 neighborhood of $p^i$, and so on so that the $k$th entry is formed using a $k\times k$ neighborhood of $p^i$, where $k\in\{1,2,\dots,10\}.$ Pixels outside of a $10 \times 10$ neighborhood of $p^i$ are not considered to be of significant impact on $p^i$. We state the results for each of these approaches in the Results and Discussion section below. 






\newpage
\subsection{Segmentation}
Segmentation via graph cuts is formulated in terms of energy minimization; more precisely as finding the optimal surface with the minimal cost. (See the graph cuts section of the Introduction for a brief overview of the theory behind graph cuts.)  Let  $F(L)$ be the energy functional to be minimized in the BK algorithm described above. 

\begin{equation}\label{F}
F(L)=\sum_i||I_i-\mu_{L_i}||_2 + \lambda\left(\sum_{\substack{\{i,j \, |\, L_i\neq L_j, \\ \ i, j \text{ are neighbors}\}}} \min\{ ||I_i-I_j||_2^{-1},1\}\right),
\end{equation}

\noindent where $I_i$ denotes the $i$th vectorized pixel  with label $L_i=1,2$  that has a predefined cluster center $\mu_1$ or  $\mu_2,$ respectively. $L$ denotes a particular labeling scheme, i.e. segmentation of the image $I$, such that $L\in\mathcal{L}$, the set of all possible labelings of the pixels in $I$. ``Neighbors'' are defined via 8-connectivity. Further, we fix $\lambda\equiv 1$ for all the calculations below and incorporate a normalizing feature for both terms, as described below.

\subsubsection*{The Data Term}
 We may interpret the first term in the functional as a data fidelity term that computes the 2-norm of the difference between each vectorized pixel and the mean of those vectorized pixels in the region it is labeled with. After preprocessing, means are derived from sample regions obtained from regions of interest (ROI's) produced by expert radiologists. The first time we run the graph cut, the two labels used are that of \textit{healthy} and \textit{tumor}, and thus $\mu_1$ is the mean value of the preprocessed pixel vectors from a sample healthy region in $I$, and $\mu_2$ is the mean value from a tumor region. The second time, we use the labels of \textit{vessel} and \textit{tumor} and their corresponding sample means. We note that the primary reason for performing the second graph cut is to obtain a more complete segmentation of the tumor regions and not to  obtain a precise location of the vessel regions, which proves to be a challenging task in and of itself.

\subsubsection*{The Perimeter Term}

We may interpret the second term in the functional as a perimeter regularization term that computes the edge weight assigned between two pixels that are in different labels. In this case, it is the 2-norm of the difference between any two vectorized pixels along a boundary. This means that perimeter penalties are charged only where the edge between two pixels is cut. Normally this term, which we may denote by $B(p,q)$, is large when pixels $p$ and $q$ are
similar (e.g. in their intensity) and $B(p,q)$ is close to zero
when the two are very different.  For example, costs may be based on local intensity gradient,
Laplacian zero-crossing, gradient direction, geometric, or other criteria \cite{boykov2001interactive}. 

\noindent One standard form \cite{boykov2001interactive} of this term   is $$B(p,q)\propto exp\left(-\frac{(I_p-I_q)^2}{2\sigma^2}\right)\cdot\frac{1}{dist(p,q)}.$$

\noindent Another perimeter term in the literature \cite{wu20173d} is $$B(p,q)=((I_p-I_q)^2+1)^{-1}+\lambda I_{sg},$$
when $p$ and $q$ are adjacent, and $B(p,q)=0$ otherwise. Here $I_{sg}$ encapsulates the tumor gradient, and $\lambda>0$. 

We propose a simpler variant of the latter $$\min\{ ||I_i-I_j||_2^{-1},1\},$$ which requires no user input or training data. Loosely speaking, it encourages cuts where the norm of the difference of the vectorized pixels is large, such as on ground truth boundaries, and discourages them where this norm is small, with a maximum penalty per cut of one. However, cuts can become too ``cheap'' and lead to grainy segmentations, which is why we introduce the following normalizing component to the data and perimeter terms in \eqref{F}.

\subsubsection*{Terminal and Edge Weight Matrices}
For the BK algorithm, we compute  the \textbf{terminal weight} matrix, which carries the weights for each pixel to the source and sink, and the \textbf{edge weight} matrix, which represents the cost of cutting between pixels.  We  let our \textit{source} and \textit{sink} each be one of $\mu_1$ or $\mu_2$, the tissue mean  vectors that  we generate from the tissue sample ROI's provided by taking the entrywise mean of the vectorized pixels in the corresponding region. We then compute the terminal weight matrix with $||I_i-\mu_1||_2$ and $||I_i-\mu_2||_2$ forming the entries in each column, $ i = 1, 2, \dots, 512^2$. We compute the edge weight matrix by first defining an 8-neighborhood of each pixel, by which we form a set of edges and the positions of their nodes, with the latter comprising the first two columns of the edge weight matrix. The weights for all edges  are then computed using the perimeter term $\min\{ ||I_i-I_j||_2^{-1},1\}$, where $I_i$ and $I_j$ are neighbors, $i\neq j$. Note that the forward and reverse directions along an edge both carry the same weight since the graph is undirected. 

In order to fix $\lambda\equiv1$ and prevent the cost of cuts from being too low, we normalize the resulting terminal weights and edge weights by dividing each weight vector by its maximum value. For each vectorized pixel $p^i$, this results in the corresponding values in the data and perimeter terms being between 0 and 1 so that the two terms in the functional \eqref{F} are equally scaled and, intuitively, the graph cut doesn't over- or undercharge for either term as compared to their magnitudes. Otherwise, $\lambda$ would need to vary over different segmentations, which is not practical. Lastly, by using the BK algorithm twice, we obtain a 3-label segmentation for our given image in terms of \textit{healthy,} \textit{tumor,} and \textit{vessel} tissues.

\subsubsection*{Time Series Matrix Algorithm}
\textbf{input:} The paths of the 59 image folders and the slice depth\\
\textbf{output:} A matrix $M$ of vectorized pixels and the 59th image for segmenting
\begin{enumerate}[(1)]
\item Load the 59 time series image folders
\item Obtain the slice location of each image from each folder
\item Create a 3D array of the images  $512 \times 512  \times 59$
\item Set the smoothing parameter (if desired) for the images 

\item Perform the smoothing (if any) using 2D convolution
\item Reshape the time series into a $512^2 \times 59$ matrix $M$
\end{enumerate}

\subsubsection*{Segmentation Algorithm}

\textbf{input:} Time series matrix $M$ of vectorized pixels, 59th image, and ROI paths for
 each tissue type\\
\textbf{output:} Segmented 3-label image

\begin{enumerate}[(1)]
\item Load ROI's for vessel, tumor, and healthy tissues
\item Generate 8-connectivity matrix $B$ from $M$
\item Compute edge weight matrix from $B$ using perimeter term
\item Calculate the three tissue mean vectors using ROI's and $M$
\item Compute the terminal weight matrix from  $M$ using data term

\item Run BK graph cut using healthy and tumor means to get Labels1 vector

\item Find only those vectors in Labels1 labeled tumor
\item Run BK graph cut  using vessel and tumor means to get Labels2 vector
\item Apply the pre-made liver mask to the full CT image
\item Assign color scales using both sets of labels and reshape output image 
\end{enumerate}  

\subsection{Evaluation}
As a quantitative assessment of the proposed model, we utilized the ground truth expert segmentations for the data set along with three standard statistical performance measures commonly reported in the literature \cite{heimann2009comparison}, \cite{moghbel2016automatic}, \cite{taha2015metrics}, \cite{christ2017automatic}, \cite{wu20173d} and the run time.

\subsubsection*{Volumetric Overlap Error} 
To compute the volumetric overlap error (VOE), we divide the number of pixels in the intersection of a segmented tumor (S) and the ground truth (T) by the total number in the union. A score of 0 represents perfect segmentation. Note that in the case of 2D data, we use the term volume in place of area for VOE and RVD.

$$VOE(\%)=\left(1-\frac{|S \cap T|}{|S \cup T|}\right)\times 100$$
\vspace{-2em}
\subsubsection*{Relative Volume Difference}
The relative volume difference (RVD) represents a measure of over- or under-segmentation of a tumor region (positive being over and negative being under). Yet, it is not a standalone indicator of performance since if the segmented tumor (S) has the same volume as the 
ground truth (T), a perfect segmentation value of 0 would be reported even if the two regions did not overlap. 

$$RVD(\%)=\left(\frac{|S|}{|T|}-1\right)\times 100$$
\vspace{-2em}
\subsubsection*{Dice Similarity Coefficient}
The overall performance of the segmentation is given by the Dice similarity coefficient (DSC), where a score of 1 represents a perfect segmentation.

$$DSC=\frac{2|S\cap T|}{|S|+|T|}$$
\vspace{-2em}
\section{Results and Discussion}
\enlargethispage{2\baselineskip}
Figures \ref{fig1} and \ref{fig2} below present the segmentations using the proposed method and the ground truth segmentations for each patient. In all images, yellow represents tumor tissue, green represents vessel tissue, and blue indicates healthy tissue.

\begin{figure}[H]
\centering
  \begin{minipage}[b]{0.49\textwidth}
     \scalebox{.30}{\includegraphics{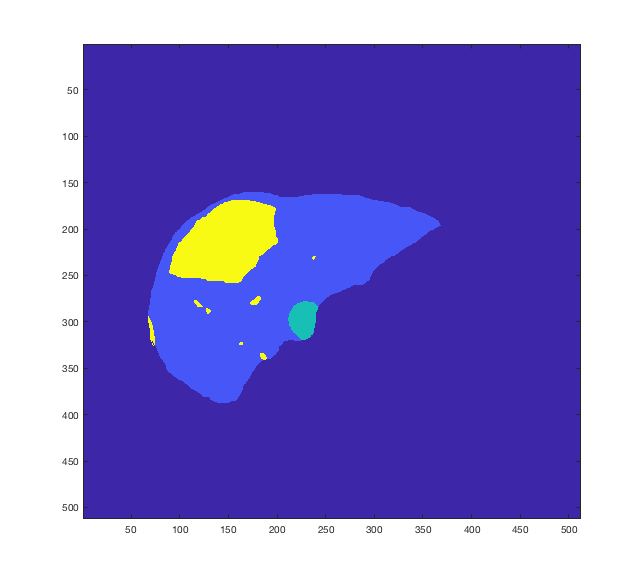}}
  \end{minipage}
  \begin{minipage}[b]{0.49\textwidth}
   \scalebox{.5}{\includegraphics[width=13.5cm,height=12.3cm]{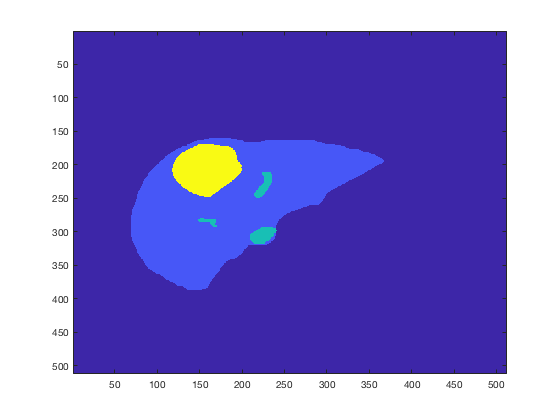}}
  \end{minipage}
  \begin{minipage}[b]{0.49\textwidth}
     \scalebox{.30}{\includegraphics{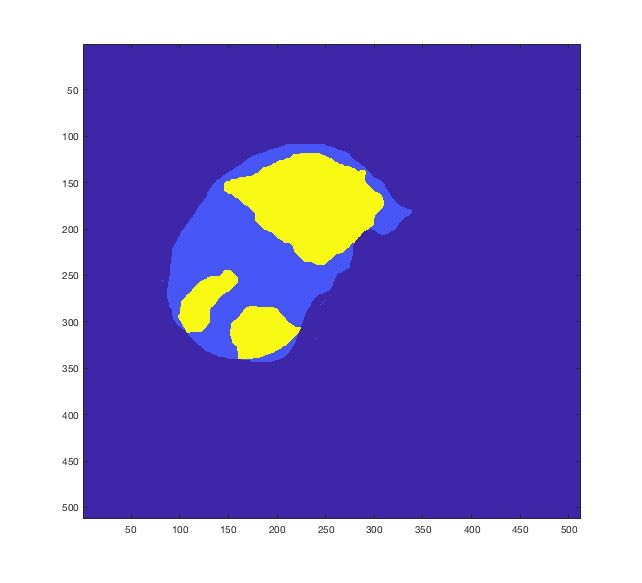}}
  \end{minipage}
  \begin{minipage}[b]{0.49\textwidth}
   \scalebox{.30}{\includegraphics{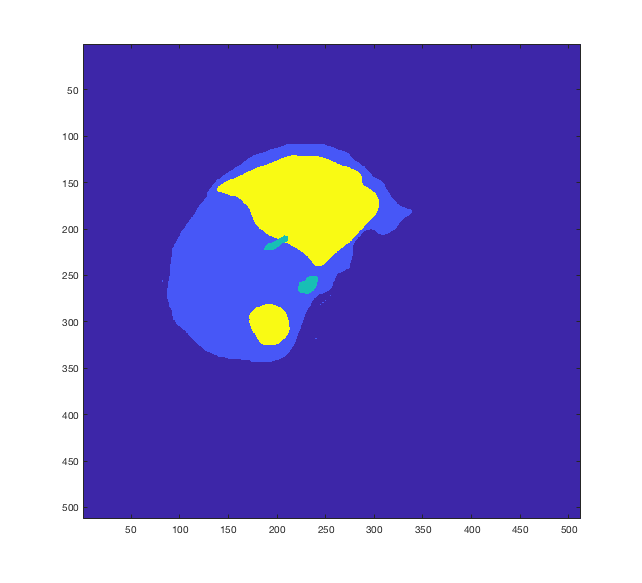}}
  \end{minipage}
    \begin{minipage}[b]{0.49\textwidth}
     \scalebox{.30}{\includegraphics{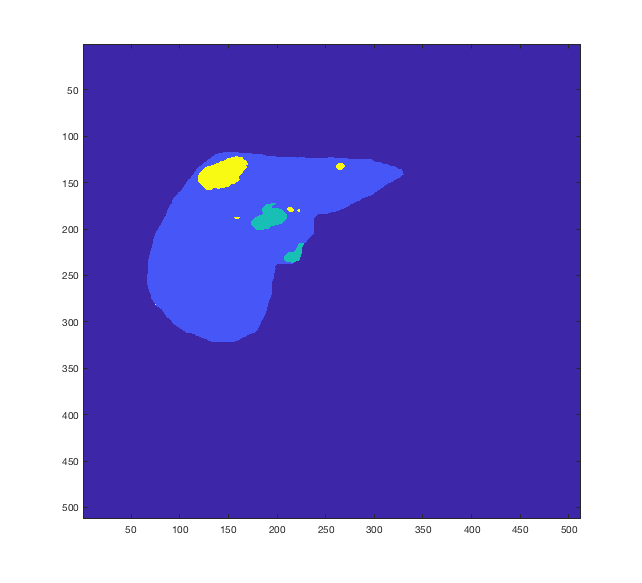}}
  \end{minipage}
  \begin{minipage}[b]{0.49\textwidth}
   \scalebox{.30}{\includegraphics{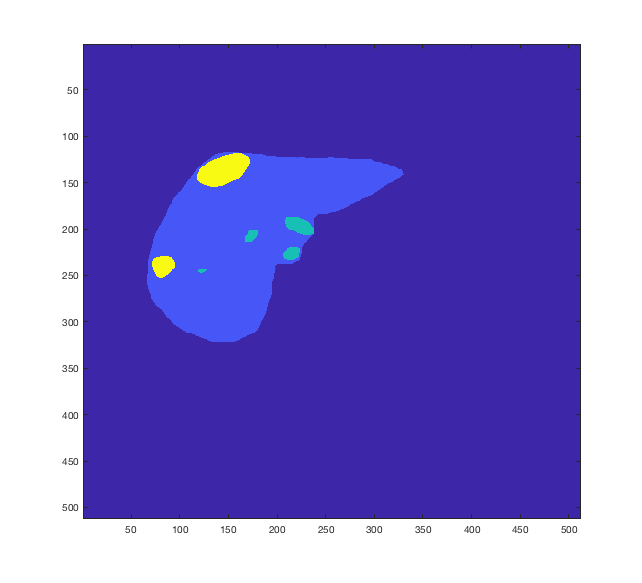}}
  \end{minipage}
\caption{The segmented image on the left and ground truth image on the right for the first five tumors using the proposed method. Yellow is tumor and green is vessel tissue. } \label{fig1}
\end{figure}

\begin{figure}[H]
\centering
  \begin{minipage}[b]{0.49\textwidth}
     \scalebox{.30}{\includegraphics{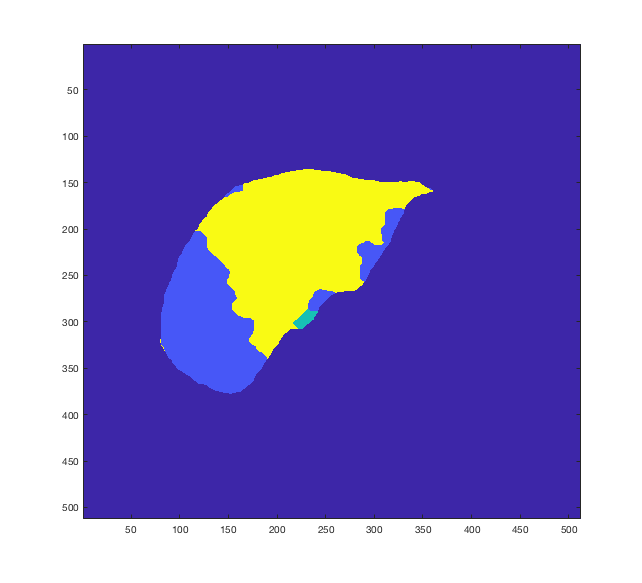}}
  \end{minipage}
  \begin{minipage}[b]{0.49\textwidth}
   \scalebox{.5}{\includegraphics[width=13.5cm,height=12.3cm]{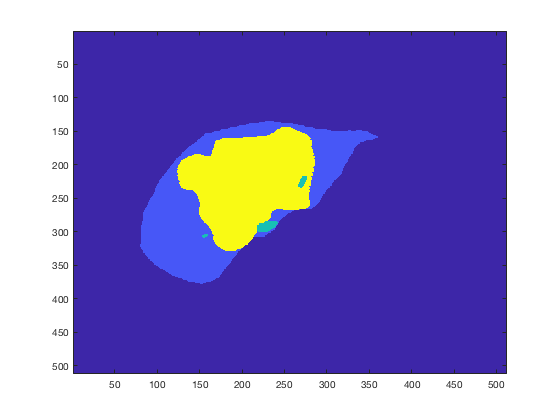}}
  \end{minipage}
  \begin{minipage}[b]{0.49\textwidth}
     \scalebox{.30}{\includegraphics{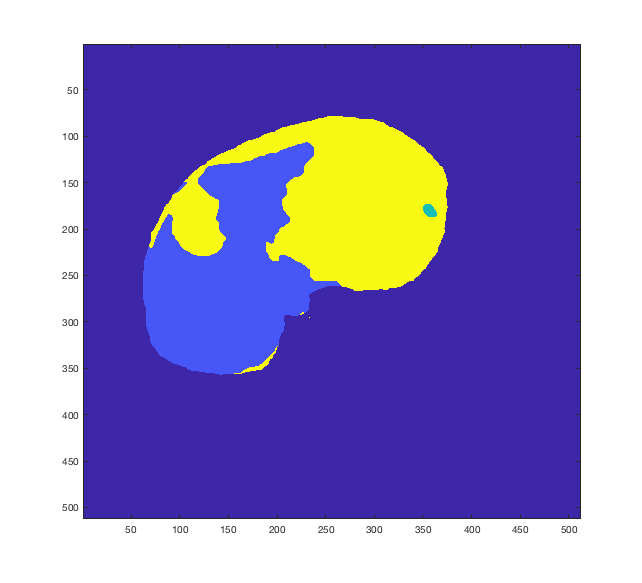}}
  \end{minipage}
  \begin{minipage}[b]{0.49\textwidth}
   \scalebox{.30}{\includegraphics{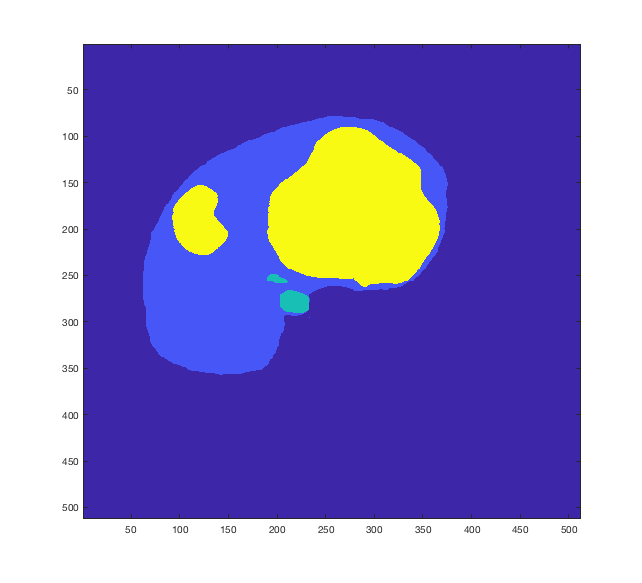}}
  \end{minipage}
    \begin{minipage}[b]{0.49\textwidth}
     \scalebox{.30}{\includegraphics{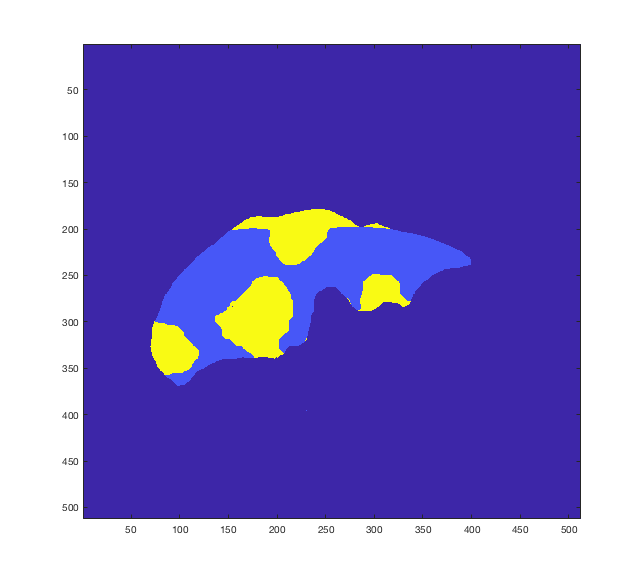}}
  \end{minipage}
  \begin{minipage}[b]{0.49\textwidth}
   \scalebox{.5}{\includegraphics[width=13.5cm,height=12.3cm]{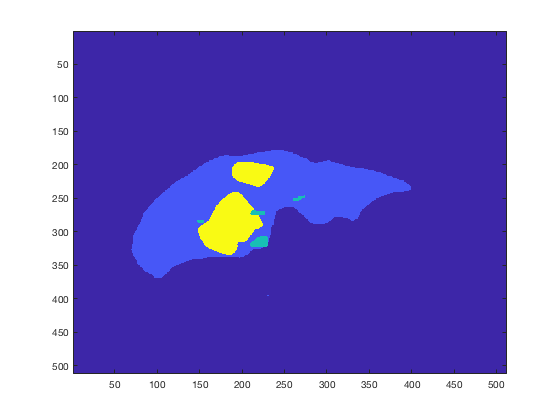}}
  \end{minipage}
\caption{The segmented image on the left and ground truth image on the right for the second five tumors using the proposed method. Yellow is tumor and green is vessel tissue. }\label{fig2}
\end{figure}

Table 1. presents the statistical comparison of the proposed method with other methods from the literature, including both automatic and semi-automatic. Data sets for the other methods shown include several different clinical CT sets, 3Dircadb, and LTSC, which came from the Medical Image Computing and Computer Assisted Intervention Society's (MICCAI) Liver Tumor Segmentation Challenge 2008. Note that while most of the other methods shown use 3D data sets and/or 3D approaches, comparing 2D results with 3D results is standard practice \cite{mourya20182D},  \cite{beichel2012liver},  \cite{moghbel2016automatic},  \cite{li2012new}. See \cite{ray2008comparison} and \cite{evans2006automatic} for a discussion of this comparison. 

When compared to the expert provided ground truth segmentations for the 10 tumors in  the clinical CT dataset, the proposed method obtains a VOE of 36.7\%, RVD of 35.1\%, DSC of .77, and mean runtime of 1.25 min/slice. We observe that using the time series feature vector significantly outperforms that of using the convolution-based feature vector, described in the preprocessing section above, and that of using only a median filter and scalar pixel intensities, with  VOE 53.1\% and 61.5\%, respectively, and DSC .58 and .48, respectively. While the RVD score is higher for the proposed method than either of these alternatives (31.6\% and 23.5\%, respectively), it should be recalled that segmentation methods can achieve a good score in this area and yet still be very inaccurate as they only have to segment the same volume as that of the ground truth but may overlap little or none at all.

Compared to the other methods shown in Table 1., the proposed method scores well in the DSC metric (although many methods do not report this value), which measures the overall performance of the algorithm. It is also comparable to the DSC score of .83 for manual segmentation on the MIDAS dataset of 10 tumors \cite{moghbel2016automatic}. It performs well in terms of runtime, being faster per slice than five of the nine other methods reporting runtime. In this regard, it is a viable alternative for expert raters since the average manual segmentation time often cited is 4.2 minutes per tumor \cite{hame2012semi}. While some methods' training time can be up to 6 hours on average  \cite{kadoury2015metastatic}, it is important to note that no training process or time is required for the proposed method; only the three sample regions of each tissue type need to be provided as ROI's to compute the tissue sample means used in \eqref{F}. Another important advantage is that the algorithm is relatively simple  to implement computationally when compared to most of the other methods shown. The results achieved are noteworthy in this way even though they are not the highest overall scores.

\begin{figure} [H]
\vspace{-5em}

    \hspace{-5.3em}\scalebox{.84}{\includegraphics{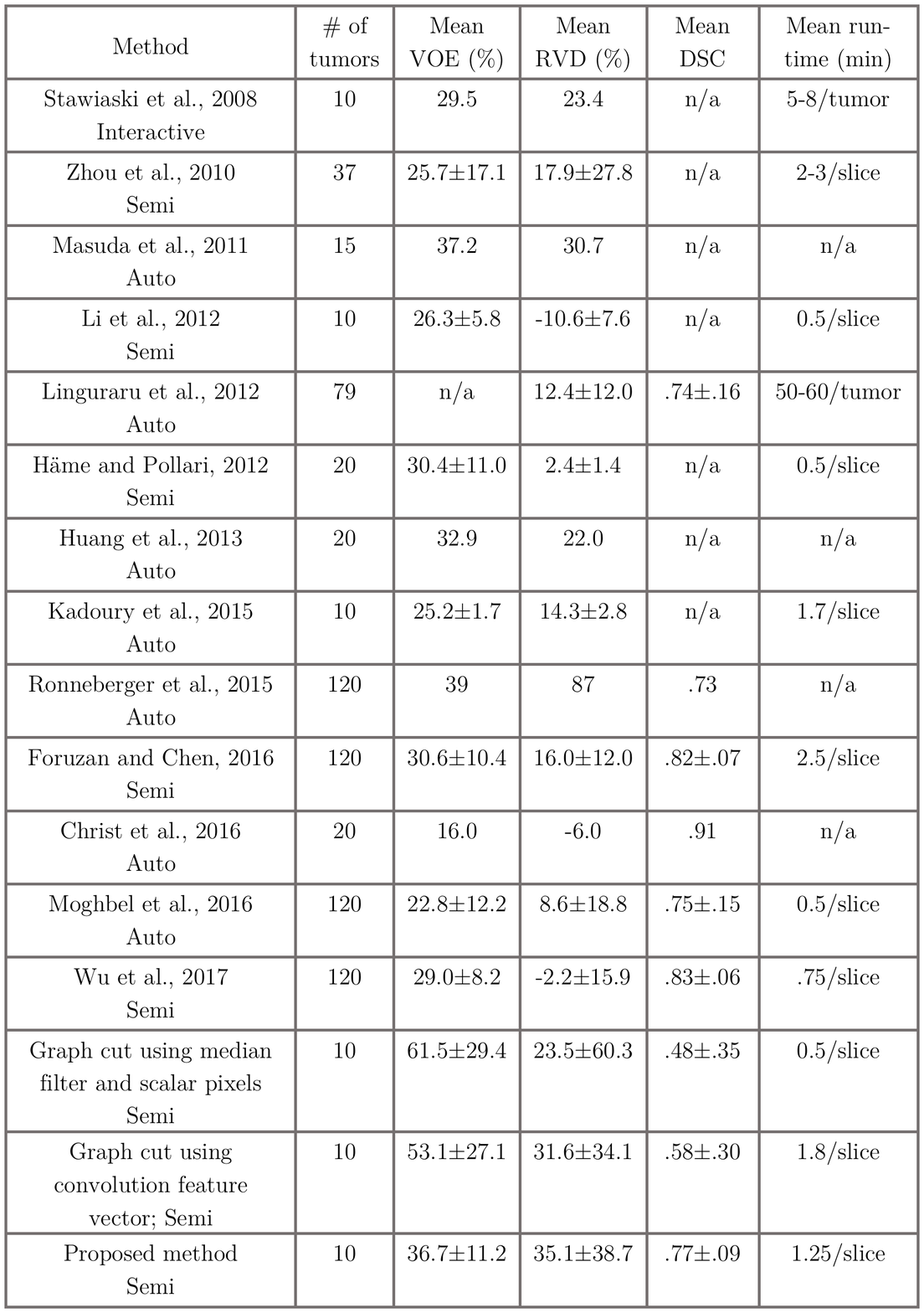}}
      \vspace{-6.5em}
      \caption*{Table 1. shows the statistical measures for the proposed method versus other methods      
        \cite{zhou2010liver},  \cite{stawiaski2008interactive},
      \cite{masuda2011liver},
      \cite{li2012new},    \cite{linguraru2012tumor}, \cite{hame2012semi}, \cite{huang2013liver},
      \cite{kadoury2015metastatic},
       \cite{ronneberger2015u},
      \cite{foruzan2016improved},
      \cite{christ2017automatic}, \cite{moghbel2016automatic},  \cite{wu20173d}.}  \label{table}
\end{figure}

Looking at the four other methods in Table 1{.} that incorporate graph cuts (GC) \cite{stawiaski2008interactive},
       \cite{linguraru2012tumor}, \cite{moghbel2016automatic},  and \cite{wu20173d}, we see similar results as just described. The proposed method's DSC of .77 is higher than two of the other three GC methods reporting; the runtime is comparable to all four; and the VOE of 36.7\% is only 7\% higher than two of the other GC approaches. However, the RVD of 35.1\% is significantly higher than the other GC methods. 
       
       The contributions of the proposed method are as follows: 1) The time series data is incorporated into a feature vector to make the BK algorithm more robust to weak boundaries and noise. 2) A simplified perimeter term is used together with a normalization on both the data and perimeter terms, which scales both terms and removes the need for optimizing $\lambda$ in the energy functional. 3) A fast runtime and relatively accurate segmentation are achieved for an algorithm of  low computational complexity that requires no training process. 
       
       Several limitations of the method are that it often misses small tumors (which has a large effect on the VOE since it is a relative measure), but oversegments in other areas (which results in the RVD being higher than desired). Both of these are most likely due to the low contrast or weak boundaries of the tumors, which pose a significant challenge in liver tumor segmentation, in general. To improve these issues, additional preprocessing could be performed, such as registration of the 59 images in the series to obtain a more precise time series representation for each pixel or image contrast enhancement techniques.  Also, the second graph cut did not perform well at locating the vessel tissues. While this was secondary to tumor segmentation in this case, it is deserving of further investigation.

\section{Conclusions}

In this paper, we propose a semi-automatic method of liver CT scan segmentation that incorporates the time series data in a novel way by creating a feature vector for each pixel using the 59 intensity values over time. We use a simplified perimeter cost term and normalize the data and perimeter terms in the functional to execute the graph cut without having to optimize the scaling parameter $\lambda$ in \eqref{F}.  Predetermined tissue means are provided in the functional that are computed from sample  regions of interest (ROI's) produced by experts, which we use to perform a graph cut twice using the BK max-flow algorithm. We first separate healthy and tumor tissues and then vessel and tumor tissues. The proposed method yields a relatively high degree of accuracy (e.g., a mean DSC of .77) for a short runtime (1.25 min./slice) and an algorithm of comparatively low computational complexity; e.g., there is no training process required.
We shall implement additional features to address the higher than desired RVD value and enhance the
ability of the proposed method to accurately detect small tumors. Testing the method on additional datasets would also be beneficial.

\section{Acknowledgements} The authors wish to sincerely thank Matthew Sottile, Noddle.io, for his highly valuable insights and contributions in reviewing this manuscript. His recommendations greatly enhanced the proposed segmentation method.


      


\bibliographystyle{plain}
\bibliography{3_label_convolution}



\end{document}